# Analyzing Linguistic Complexity and Scientific Impact

Chao Lu[1], Yi Bu[2,3], Xianlei Dong[4], Jie Wang[5], Ying Ding[2,6], Vincent Larivière[7], Cassidy R. Sugimoto[2], Logan Paul[2], Chengzhi Zhang[1]*

*1. School of Economics and Management, Nanjing University of Science and Technology, Nanjing, Jiangsu, China*
*2. School of Informatics, Computing, and Engineering, Indiana University, Bloomington, Indiana, U.S.A.*
*3. Center for Complex Networks and Systems Research, School of Informatics, Computing, and Engineering, Indiana University, Bloomington, Indiana, U.S.A.*
*4. School of Management Science and Engineering, Shandong Normal University, Jinan, Shandong, China*
*5. School of Information Management, Nanjing University, Nanjing, Jiangsu, China*
*6. School of Information Management, Wuhan University, Wuhan, Hubei, China*
*7. École de bibliothéconomie et des sciences de l'information, Université de Montréal, Montréal, Québec, Canada*

**Corresponding author: Chengzhi Zhang**, Email: zhangcz@njust.edu.cn

# Analyzing Linguistic Complexity and Scientific Impact

**Abstract**: The number of publications and the number of citations received have become the most common indicators of scholarly success. In this context, scientific writing increasingly plays an important role in scholars' scientific careers. To understand the relationship between scientific writing and scientific impact, this paper selected 12 variables of linguistic complexity as a proxy for depicting scientific writing. We then analyzed these features from 36,400 full-text Biology articles and 1,797 full-text Psychology articles. These features were compared to the scientific impact of articles, grouped into high, medium, and low categories. The results suggested no practical significant relationship between linguistic complexity and citation strata in either discipline. This suggests that textual complexity plays little role in scientific impact in our data sets.

## 1. INTRODUCTION

The success of scholars can be assessed with the aid of several indicators (e.g., high-impact publications, distinguished positions, prizes, etc.). Among these, high-impact publications have become one of the most important criteria, and several scholars have attempted to understand the factors that affect the impact of scholarly works (e.g., Amjad *et al.*, 2017; Onodera & Yoshikane, 2015). Wang, Song & Barabási (2013) found that fitness (accounting for the perceived novelty and importance of a discovery) plays a vital role in affecting the long-term impact of a work. Amjad *et al.* (2017) found that there is a positive correlation between collaboration with advanced researchers and more citation counts of the publications in a domain. Other variables have also been shown to have a correlation with citation counts, such as publication venues and review cycles (Larivière & Gingras, 2010; Shen *et al.*, 2015; Onodera & Yoshikane, 2015; Tang, Shapira, & Youtie, 2015; Waltman, 2016) as well as collaboration (Larivière, Gingras, Sugimoto, & Tsou, 2015; Wu, Wang, & Evans, 2019; Wuchty, Jones, & Uzzi, 2007; Zhang, Bu, Ding, & Xu, 2018).

The growth of big data provides new opportunities and challenges for the field of bibliometrics (Ding *et al.*, 2014). The advent of digital publishing, which has led to an abundance of structured full-text scholarly documents, allows for opportunities that did not exist when Garfield developed the Science Citation Index. Several publishers provide the full-text of open-access papers (e.g., *PLoS*[1]), which can be used to enrich existing studies (Ding & Stirling, 2016). The availability of both full-text and metadata has allowed for the combination of computation linguistics and citation analysis (e.g., Bertin, Atanassova, Gingras, & Larivière, 2016; Ding *et al.*, 2013; McKeown *et al.*, 2016; Teufel, 2000; Wan & Liu, 2014). Several studies have examined the relationship between writing features and scientific impact, focusing on *inter alia*, title length, abstract length, keywords selection, and figure usage within the text (e.g., Didegah & Thelwall, 2013; Lee, West, & Howe 2018; Moat & Preis, 2015; Uddin & Khan, 2016). However, these studies have emphasized external features or descriptive attributes rather than the internal structure and writing. To address this gap, we examined the relationship between linguistic complexity and scientific impact, following the framework developed in Lu *et al.* (2019).

Linguistic complexity generally takes two forms—syntactic and lexical (Levinson, 2007; Lu *et al.*, 2019; Nolan, 2013)—by which the variety and sophistication of forms in language production can be quantitatively measured. Taking advantage of the recent development in computational linguistics techniques and the availability of the full-text of scholarly documents, this paper investigated the relationship between the linguistic complexity of scholarly publications and their scientific impact. By understanding the relationship between them, scholars can achieve a better knowledge of the role of linguistic complexity in scientific writing and metrics of success.

---

[1] https://www.plos.org/

To analyze this relationship, we divide all publications into three groups based on their normalized number of citations—namely, high-, medium-, and low-impact groups in two selected disciplines. We selected 12 variables to represent the linguistic complexity of a publication and examined whether there were any differences among these strata.

## 2. RELATED WORK

### 2.1. Linguistic complexity in scientific writing

Scientific writing is a fundamental part of the scientific process, allowing for the communication and exchange of ideas among scholars (Hyland, 2004). As Montgomery noted: "There are no boundaries, no walls, between the doing of science and the communication of it; communicating is the doing of science" (c.f. Cronin, 2005). Despite disciplinary differences in discourse (e.g., Demarest & Sugimoto, 2015), there are several aspects of scientific writing that make it distinct from other forms of communication. First, many disciplines have adopted a structured format to the text, such as the IMRD (i.e., Introduction, Methods, Results, and Discussion) format (Gastel & Day, 2016). Furthermore, there are several inherent features of scientific writing (e.g., Biber & Gray, 2010; Gastel & Day, 2016; Gopen & Swan, 1990) that make it distinct from spoken language. Many of these distinctions are due to the linguistic complexity of scientific writing (Biber & Gray, 2010; Fang, 2005; Gray, 2011). Specifically, scientific writing contains more noun phrases embedded with modifiers, which makes the sentences more structurally compressed and abstracted. Similarly, Snow (2010) reported that scientific writing is exceptionally concise and contains a high density of information. Likewise, scientific articles in 64 journals across several disciplines (e.g., humanities, social sciences, natural sciences, and engineering) have been found to possess similar syntactic complexity: clauses are frequently adopted; a variety of modifier structures have been observed across disciplines, the average sentence length is 24.9 words with a standard deviation of 11.6, and usually 126.7 words comprise a paragraph (Gray, 2011).

## 2.2. Computational Linguistics and Bibliometrics

Studies in computational linguistics have utilized various natural language processing technologies to develop indicators that investigate textual data quantitatively (e.g., Brants, 2000; Brown *et al.*, 1993). For example, the CAF indicators—i.e., Complexity, Accuracy, and Fluency—are among the most adopted measurements in English proficiency, especially for writing (Ellis & Yuan, 2004). Complexity has various advantages over accuracy and fluency in measuring English scientific writing style (Lu *et al.*, 2019). It comprises two aspects: syntactic and lexical complexity (Ferris, 1994a; Kormos, 2011; Ojima, 2006). Syntactic complexity consists of quantitative variables in three subgroups: sentence length, sentence complexity, and "other" (Lu *et al.*, 2019). Lexical complexity is made of lexical diversity, lexical density, and lexical sophistication (Vajjala & Meurers, 2012). They have been used in authorship attribution identification (Holmes, 1994), readability classification (Vajjala & Meurers, 2012), and gender identification in scientific articles (Bergsma, Post, & Yarowsky, 2012). Our previous work (Lu *et al.*, 2019) used this metric to investigate the difference between native and non-native English writers in scientific writing.

## 2.3. Scientific writing and scientific impact

A survey of more than 200 members of the American Psychological Society found that one of the chief attributes of highly cited work was "Quality of Presentation" (Sterberg

& Gordeeva, 1996). Since this study, there have been several attempts to identify, quantitatively, the relationship between text and scientific impact. Starting with the title, studies have shown a correlation between shorter titles and increased numbers of citations (Letchford, Moat & Preis, 2015; Sienkiewicz & Altmann, 2016). However, there is some ambiguity in this: Didegah & Thelwall (2013) found no relationship between the length of titles and citation counts of papers in Biology, Chemistry, and Social Science. Furthermore, Fox & Burns (2015) found that different types of article titles also play a role—papers with broader and more general titles receive more citations than organism-specific ones. This may demonstrate a difference in topicality, e.g., a relationship was demonstrated between selection of keywords and paper citation counts in the field of obesity (Uddin & Khan, 2016).

Abstracts have also been investigated, with associations found between length of (Didegah & Thelwall, 2013; Sienkiewicz & Altmann, 2016) and complexity (Sienkiewicz & Altmann, 2016) of abstracts and subsequent citation impact. Using the Flesch readability score, Gazni (2011) found that abstracts of high-impact articles tend to be more difficult to read than those of low-impact articles.

Studies have also begun to investigate within the text. Lee *et al.* (2018) found that higher-impact articles contain more figures and fewer tables (based on an analysis of more than 650,000 PubMed articles). These findings, however, were not confirmed by Haslam and colleagues, who showed that, in Psychology, the relationship between scientific impact and linguistic features of articles, such as title and article length, and numbers of figures and tables, are quite weak (Haslam *et al.*, 2008; Haslam & Koval, 2010).

The state-of-the-art suggests several areas of ambiguity and potential disciplinary differences in the relationship between text and scientific impact. Furthermore, there is a dearth of studies that utilize the full-text of scientific documents. The focus on metadata or descriptive attributes does not reflect the true linguistic features of writing. Therefore, we propose an analysis of the full-text features across two disciplines to advance our understanding of the relationship between scientific writing and scientific impact.

## 3. METHODS

The roadmap for this study is shown in Figure 1. First, the data set for the study was introduced. Next, the linguistic features were calculated from two perspectives (i.e., syntactic and lexical complexity). Using the citation data from Web of Science (WoS), we categorized papers into three groups. Finally, we analyzed the linguistic complexity among the three groups.

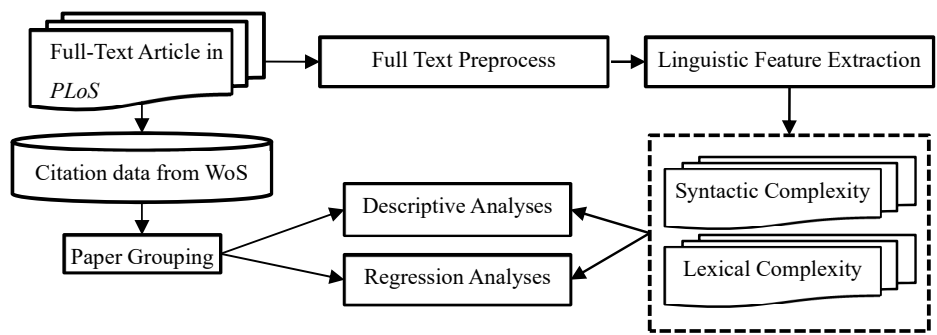

**Figure 1. Road map for this study.**

*3.1. Data*

*PLoS* is one of the largest and most high-impact open-access publishers in the world. Furthermore, they are one of the only large-scale publishers to make fully available well-structured full-text of their articles. Therefore, we selected *PLoS* as our data source. We initially collected 172,662 full-text articles published from 2003 to 2015 in the *PLoS* journal family, a set of peer-reviewed journals covering various disciplines. To mitigate the potential effect caused by disciplinary differences, we selected *research articles* from two disciplines: Biology and Psychology. Biology is the dominant high-impact discipline in *PLoS* journals while Psychology is the non-natural science subject with the highest number of publications. Given that *PLoS* assigns multiple tags to each article, the disciplinary attribute of the articles is not precise. Thus, we initially obtained two candidate article pools for the two subjects. If one article was assigned the tag Biology, it will run into our Biology candidate article pool; if one is assigned the Psychology subject tag, it will be in our Psychology candidate article pool. Our sampling frame thus consisted of 49,211 full-text articles (and associated metadata) in the Biology candidate article pool and 1,975 in the Psychology candidate article pool. At this stage, one article could be in both pools; these articles were removed in the following research stages.

After this initial filtering, we matched the PLOS records with Web of Science to obtain more fine-grained disciplinary classification data, drawing on the NSF (National Science Foundation) subject classification. From the initial Biology article pool, 6,617 articles could not be matched with records in our full WoS data set and were thus removed. The remaining 42,594 matched articles were associated with 114 domains based on the NSF classification information (shown in Table 1). To obtain more precise normalized citation data with the NSF subject classification system, we removed all of the matched publications that were identified as non-Biology papers. This left 36,400 full-text articles identified by both the *PLoS* and NSF classification systems as Biology articles. Similarly, we removed 178 articles that could not be correctly matched in WoS from our initial Psychology article pool. The remaining 1,797 articles were all identified by both the *PLoS* and the NSF classification system as Psychology articles, as shown in Table 2. Table 3 shows the associated journal information. Most articles were published in *PLoS One*.

We then retrieved the numbers of citations for all of the articles in both data sets. We calculated the field-based citation normalization using the publication year of each article and their domain information from the NSF classification.

**Table 1. The number of publications by domain in the Biology data set.**

| Domain | Freq. | Domain | Freq. |
|---|---|---|---|
| Biochemistry & Molecular Biology | 9,074 | Physiology | 245 |
| Genetics & Heredity | 4,571 | Miscellaneous Biology | 148 |
| Immunology | 3,097 | General Biomedical Research | 145 |
| Neurology & Neurosurgery | 3,067 | General Zoology | 143 |
| Microbiology | 2,880 | Oceanography & Limnology | 130 |
| Cancer | 2,062 | Miscellaneous Zoology | 113 |
| Parasitology | 1,637 | Agriculture & Food Science | 106 |
| Ecology | 1,403 | Environmental & Occupational Health | 100 |
| Virology | 1,029 | Anthropology and Archaeology | 94 |
| Cellular Biology Cytology & Histology | 1,011 | Environmental Science | 90 |
| Marine Biology & Hydrobiology | 892 | Dentistry | 84 |
| Botany | 877 | Nutrition & Dietetic | 84 |
| Endocrinology | 658 | Radiology & Nuclear Medicine | 83 |
| Cardiovascular System | 612 | Probability & Statistics | 75 |
| General Biology | 575 | Arthritis & Rheumatology | 75 |
| Pharmacology | 504 | Biophysics | 47 |
| Entomology | 400 | Miscellaneous Biomedical Research | 43 |
| Embryology | 246 | | |

**Table 2. The number of publications by domain in the Psychology data set.**

| Domain | Freq. |
|---|---|
| Experimental Psychology | 705 |
| Behavioral Science & Complementary Psychology | 522 |
| Social Psychology | 232 |
| Developmental & Child Psychology | 184 |
| Clinical Psychology | 95 |
| Miscellaneous Psychology | 47 |
| General Psychology | 12 |

**Table 3. The number of publications by journal in the biology and psychology data sets.**

| Biology | | Psychology | |
|---|---|---|---|
| **Journal** | **# (Ratio)** | **Journal** | **# (Ratio)** |
| *PLoS One* | 32,758 (0.90) | *PLoS One* | 1797 (1.00) |
| *PLoS Genetics* | 1,674 (0.05) | | |
| *PLoS Pathogens* | 1,469 (0.04) | | |
| *PLoS Biology* | 499 (0.01) | | |

### 3.2. Data Processing

First, full-text articles (XML format) were preprocessed with Python scripts. We employed NLTK[2] (Bird, Klein, & Loper, 2009) to extract all text within the tag <p> from the full-text with *re* and *xml* and then removed the remaining tags and tokenized sentences when abbreviations were replaced by their complete forms, e.g., "*et al.*" Following this, to calculate measurements for linguistic features, Stanford Parser (Dan & Christopher, 2003) was applied for part-of-speech (POS) tagging. Tregex[3] was used to extract clauses according to Lu (2010). Finally, when calculating measures of lexical complexity, we merged the POS tags given by Tree Bank. For instance, "NN" and "NNS" both counted as nouns.

### 3.3. Variables

As mentioned above, we aimed to explore the relationship between linguistic complexity and scientific impact (measured by number of citations). Normalized citations are used to mitigate the possible effects caused by different periods of citation history (Radicchi, Fortunato, & Castellano, 2008; Schubert & Braun, 1986). The variable is calculated as follows:

$$NC_i = \frac{TC_i^{PY}}{ADC_{PY}}$$

where $NC_i$ is the normalized number of citations for paper *i*, which was published in the year $PY$ in our data set. $TC_i^{PY}$ represents the total number of citations that the paper

---

[2] http://www.nltk.org/: This library integrates all kinds of natural language processing tools and datasets and provides easy-to-use interfaces.

3 https://nlp.stanford.edu/software/tregex.shtml

*i* received since its publication. $ADC_{PY}$ denotes the average number of citations received by of all the papers published in the same year $PY$ in the domain of paper *i*.

In regard to linguistic complexity, specific quantitative variables to measure the two aspects of complexity are presented in Figure 2 (the variables selected were adopted for this study), which were developed in a previous study (Lu *et al.*, 2019).

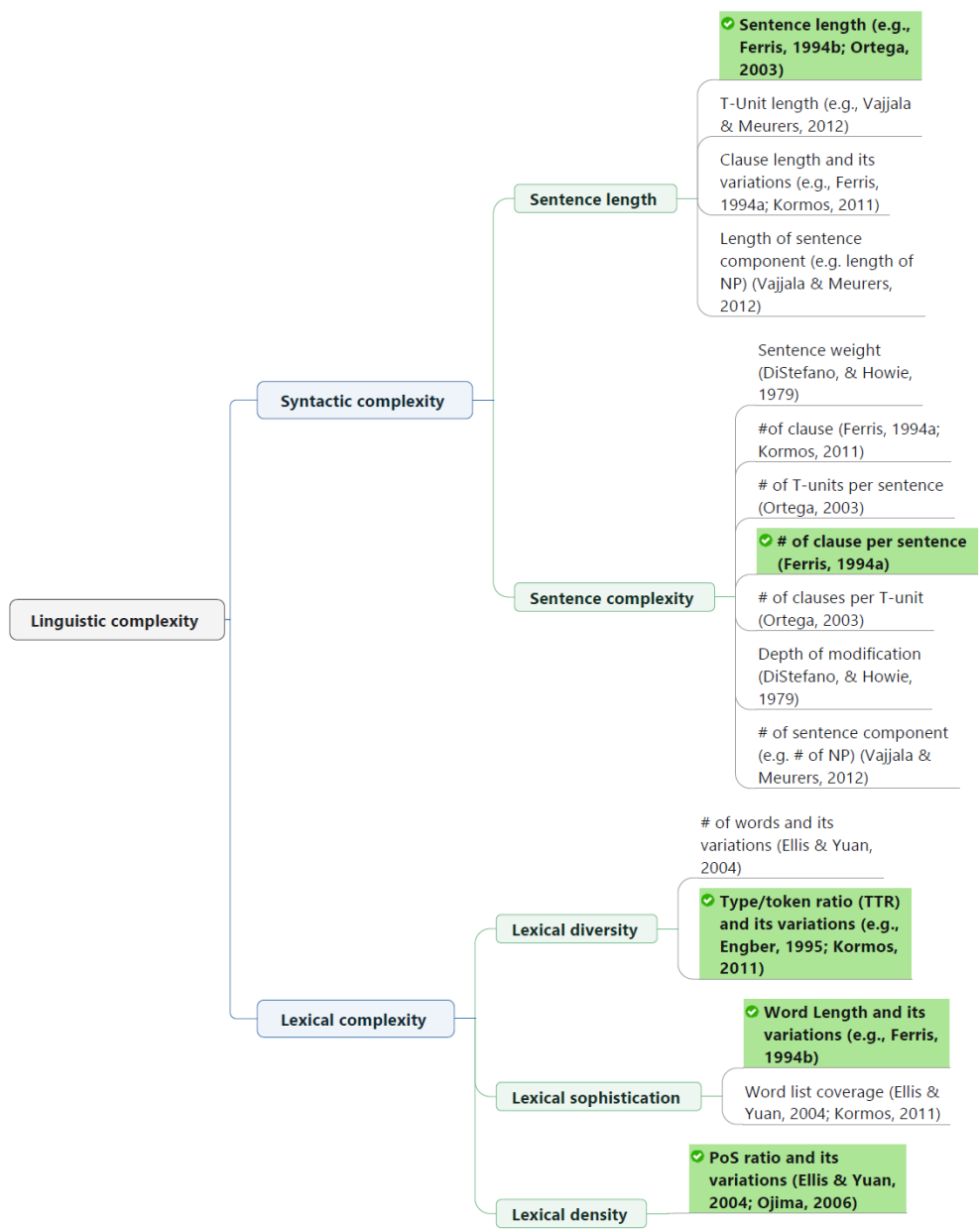

**Figure 2. Dimensions of linguistic complexity found in the literature.**

*Syntactic complexity* (called syntactic maturity or linguistic complexity) signifies "the range of forms that surface in language production and the degree of sophistication of such forms" (Ortega, 2003, p.492). Measurements of syntactic complexity can be generally categorized into three groups: sentence length, sentence complexity, and "other" measurements (Lu *et al.*, 2019). The first two groups of measurements have been widely used as indicators to assess the syntactic complexity of native and non-native English speakers in that both indicators are beneficial for identifying the language proficiency of writing (e.g., Ortega, 2003; Vajjala & Meurers, 2012). Sentence length measures the number of words in a sentence. Other similar variables include average T-unit[4]/clause length (Vajjala & Meurers, 2012). Two variables—average sentence length per article and standard deviation of sentence length per article—derived from this feature are adopted in this study following previous practices (e.g., Ferris, 1994b; Ortega, 2003). Sentence complexity measures the number of sentence phrases per sentence, including features such as the number of clauses (i.e., a structure with a subject and a finite verb[5] (Polio, 1997))/T-units/subordination (e.g., NP components) (Vajjala & Meurers, 2012). In this study, the clause ratio of each article is applied, which is calculated by dividing the number of clauses by the total number of sentences to measure sentence complexity. This indicator has been used in several previous studies (e.g., Ferris, 1994a; Lu, 2010; Polio, 1997).

*Lexical complexity* (or lexical richness) measures the richness of vocabulary in writings (Laufer & Nation, 1995). Measurements of that can be grouped into three clusters:

---

4  Shortest grammatically allowable sentences into which (writing can be split) or minimally terminable unit (Hunt, 1965).

5  A verb that has both a subject and a tense.

lexical diversity, lexical sophistication, and lexical density (Lu *et al.*, 2019; Vajjala & Meurers, 2012). Lexical diversity refers to the number of different words used in the text. This indicator is usually measured by the type/token ratio (TTR) of each article (Engber, 1995; Youmans, 1990) or the number of unique words of each article. This study uses TTR to measure lexical diversity because it is frequently used and the total number of unique words is normalized by the length of the text. Lexical sophistication signifies the degree of sophistication of lexical items (i.e., nouns, verbs, adjectives, and adverbs). It is usually measured by the average length of words in each paper or the coverage of a certain vocabulary list, which can reflect the cognitive complexity for both writers and readers (Juhasz, 2008). The first measurement is preferred to depict the lexical sophistication by calculating the average number of characters in a word (Vajjala & Meurers, 2012) for its frequent usage and reduced complexity of calculation. In this study, each kind of lexical item is collectively used to measure the lexical sophistication using the NLTK package for POS parsing, which means four variables are used. Last, lexical density is defined as the proportion of lexical items by the total number of words (Lu *et al.*, 2019). It is calculated as the ratio of lexical items to the total number of words (Lu, 2011). While lexical items provide semantic meaning, studies show that using adjectives and adverbs (collectively called modifiers) could improve the readability of the text (Vajjala & Meurers, 2012).

*3.4. Paper Grouping*

To fully consider the disparities between articles with different levels of impact as suggested in other studies (e.g., Amjad *et al.*, 2017), we also divided the articles into three categories inspired by the Essential Science Indicators[6] impact typology: high impact (G1) (top 1% most cited papers), medium impact (G2) (top 1% to top 10%), and low impact (G3) (the remaining 90%). The grouping information is shown in Table 4.

---

[6] https://clarivate.com/products/essential-science-indicators/

**Table 4. Description of Article Groups by levels of impact.**

| Subject | Group | Max of NC | Min of NC | Average of NC | # of Article | Ratio |
|---------|-------|-----------|-----------|---------------|--------------|-------|
| Biology | High Impact | 20.49 | 1.80 | 2.87 | 443 | 1% |
| | Medium Impact | 1.80 | 0.66 | 0.97 | 4,429 | 9% |
| | Low Impact | 0.66 | 0 | 0.22 | 44,304 | 90% |
| Psychology | High Impact | 19.00 | 5.46 | 7.89 | 17 | 1% |
| | Medium Impact | 5.43 | 2.13 | 3.09 | 162 | 9% |
| | Low Impact | 2.11 | 0 | 0.73 | 1618 | 90% |

# 4. RESULTS

We plotted the cumulative distribution function (CDF) for all 12 variables indicating linguistic complexity by impact group of Biology in Figure 3 with Kolmogorov-Smirnov (KS) test results.

In most features, the high-impact groups differ from low- or medium-impact groups with statistical significance. Only two lexical sophistication variables, verb length and adverb length, suggest no statistical significance between any groups by the KS test. Apart from these two variables, Figure 3 suggests that, regarding syntactical complexity, high-impact groups usually tend to present a slightly higher degree of complexity than the other two groups of articles, which might suggest that high-impact articles tend to use longer and more complex sentences than medium- and low-impact articles. Among these three features, the standard deviation of sentence length suggests it is of importance, but this has no statistical significance. That implies that in our biology data set, high-impact articles tend to use marginally longer sentences and more clauses in their manuscripts.

Similarly, most lexical complexity features suggest that high- and medium-groups differ from low-impact groups with statistical significance in our Biology data set, among which noun length, noun ratio, and adjective ratio suggest significant differences between groups. TTR (indicating lexical diversity) suggests that the high-impact group of articles tend to use slightly more vocabulary than the low-impact group but less than the medium-impact group. For lexical sophistication, noun length suggests that the high-impact group of articles uses slightly longer nouns than the medium-impact group, which is followed by the low-impact group (p≤0.01). Adjective length suggests an opposite trend: the high-impact group uses slightly shorter adjectives than the medium- and low-impact groups (p≤0.001). On lexical density, high-impact articles usually tend to use fewer nouns and verbs but more modifiers (adjectives and adverbs) than medium- and low-impact articles (p≤0.05). For instance, in adjective usage, high-impact articles use more adjectives than medium-impact articles followed by low-impact articles.

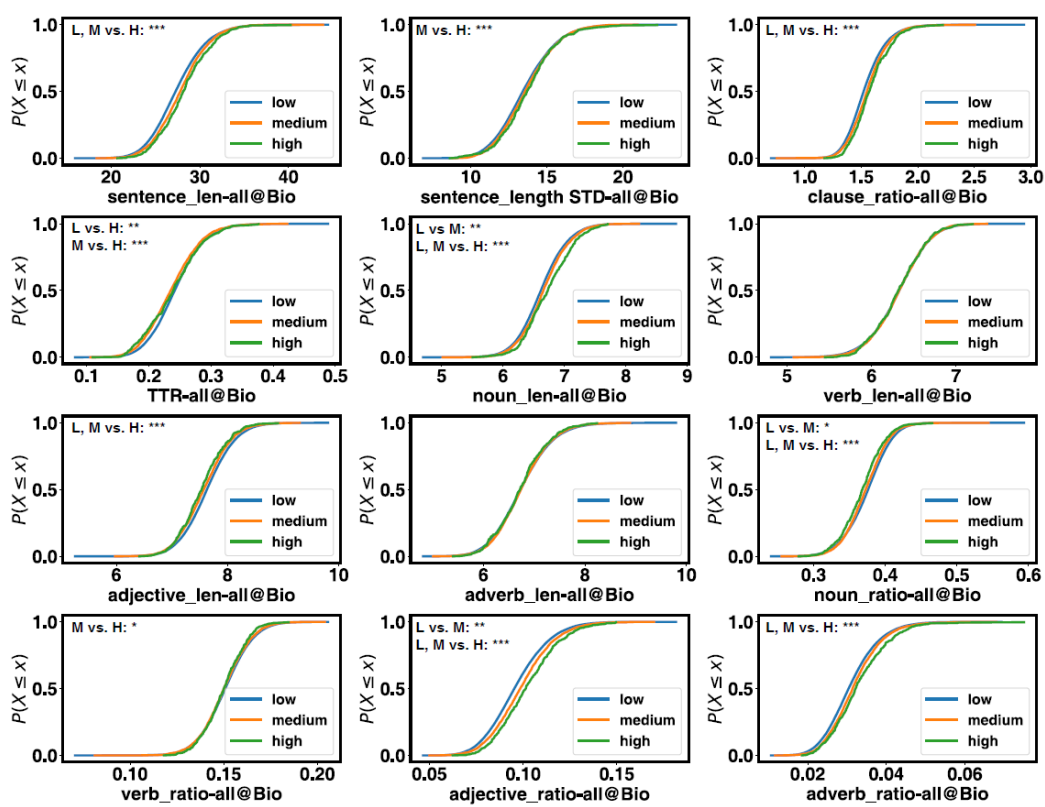

**Figure 3. CDF plots for 12 linguistic complexity features in Biology (In the plots, a dot in the plots means the probability for a certain feature is greater than x. Stars (\*) represent the statistical significance level where one star (\*) means p≤0.05, two stars (\*\*) p≤0.01, and three stars (\*\*\*) p≤0.001 using the Kolmogorov-Smirnov test. The same below).**

In regard to our Psychology data set, similar patterns emerge; however, there are no statistically significant differences among the high-, medium- and low-impact groups (shown in Figure 4 for both syntactical and lexical complexity). We attribute this observation to two possible reasons. One is that due to the limited number of articles in our Psychology data set, we could not produce smooth CDF curves for high- and medium-groups of articles of our Psychology data set. The second is that the relationship between the syntactic complexity features and scientific impact might show little practical significance in Psychology.

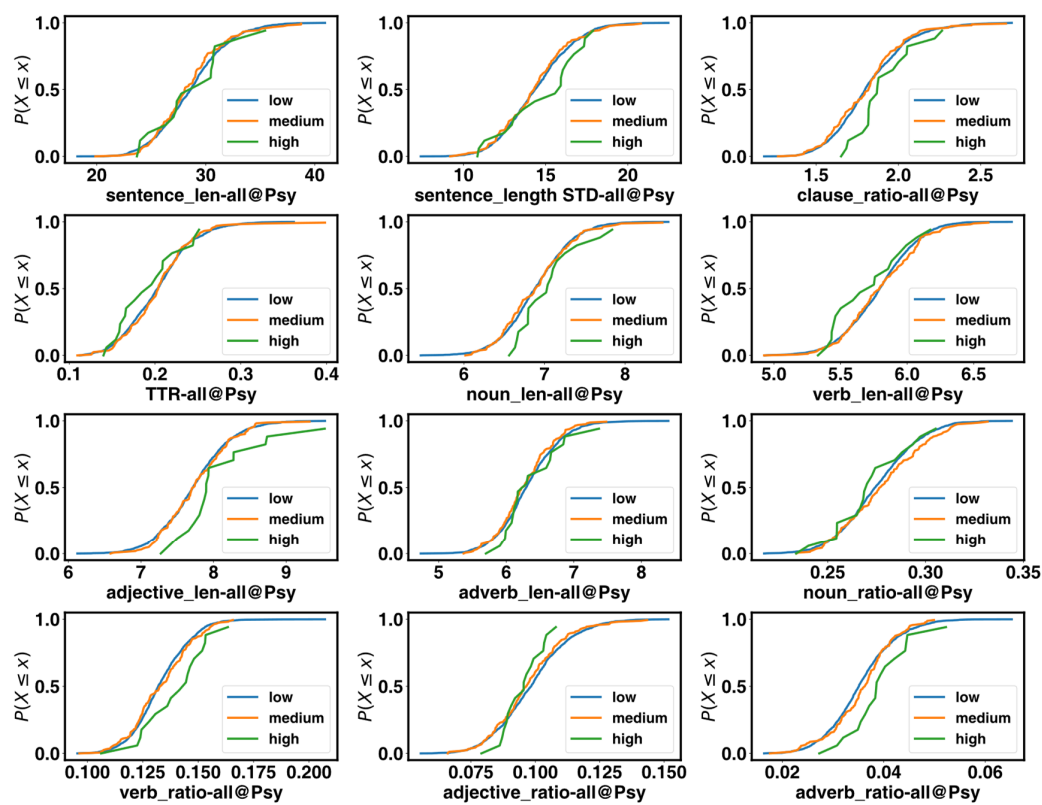

**Figure 4. CDF plots for 12 linguistic complexity features in Psychology.**

To further investigate their practical significance, we plotted the point estimations with 95% confidence intervals for each linguistic feature in Figures 5 and 6. We found that the differences between groups were marginal in each linguistic feature as suggested by the vertical axes. Specifically, the clause ratio, for instance, suggested statistical significance between the high- and medium-impact groups and between the medium- and low-impact groups, while the difference between groups is limited to no more than 0.05 according to the estimated point values in our data set. Similar observations that the practical differences are marginal between groups can be observed in those groups where statistical significance can be detected, e.g., the noun length between high- and medium-impact groups. This means that the significant results of these variables are not practical. Again, thus, it is concluded that the relationship between the linguistic complexity of a paper and its scientific impact suggests no practical significance. In addition, the confidence intervals of high-impact groups in both subjects are quite wide, indicating that the limited number of high-impact articles might reduce the precision of the point estimation results.

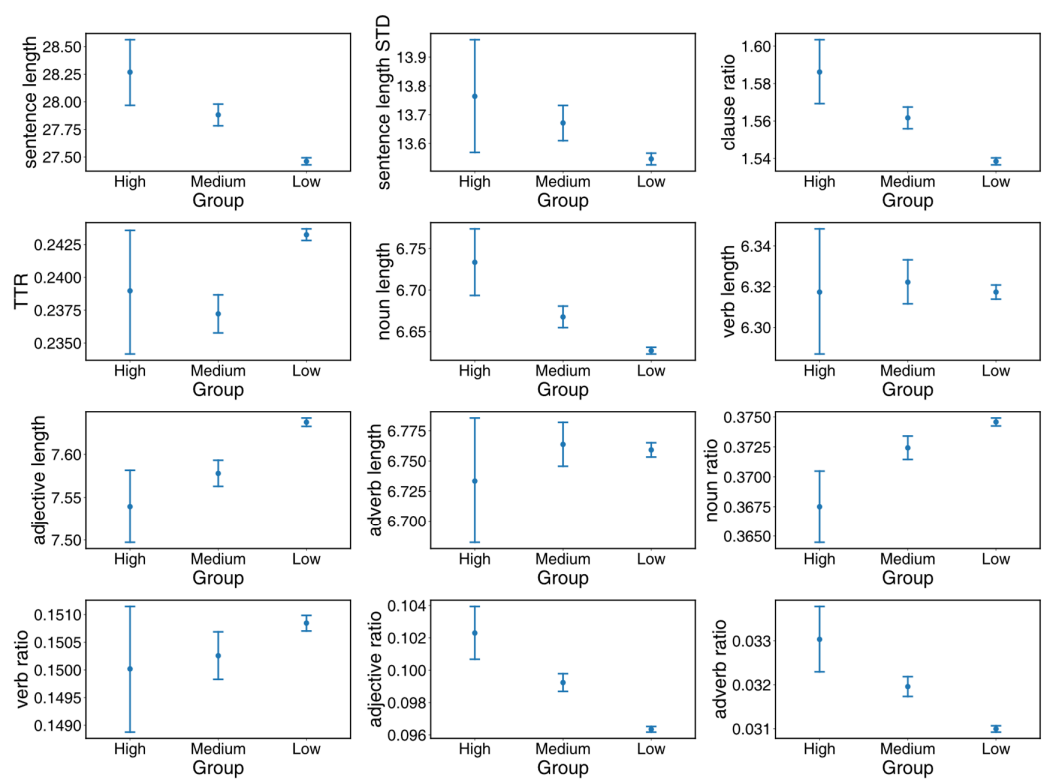

**Figure 5. Point estimation plots for 12 linguistic features by group in Biology (the error bars represent the confidence intervals using bootstrapping with 10,000 iterations. The same below).**

We also conducted regression analyses by employing both logarithm and quadratic transformations[7], but the models show very low R-squares (see details in the Appendix), indicating weak and complex relationships between linguistic complexity and scientific impact in these datasets.

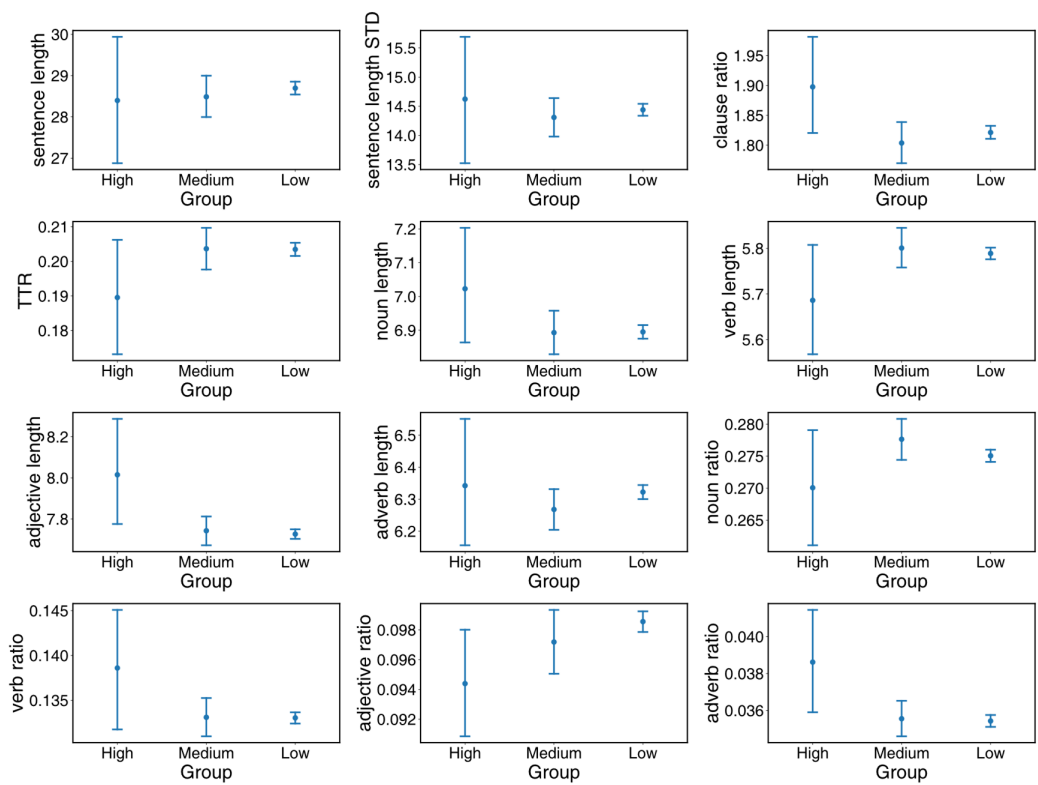

**Figure 6. Point estimation plots for 12 linguistic features by group in Psychology.**

## 5. DISCUSSION AND CONCLUSIONS

This paper combined bibliometrics and computational linguistics to better understand the relationship between scientific writing and scientific impact. The results for Biology show significantly different differences for most variables across the levels of impact (Figure 4); however, the point estimation results indicate that the differences between groups are not practically significant; in our Psychology data set, the relationship between linguistic complexity and scientific impact is even weaker (though this could also be an artifact of sample size). This suggests a weak relationship between the linguistic complexity of a paper and scientific impact, which is reinforced by the results in our regression analyses.

---

[7] We followed the experience from Didegah and Thelwall (2013) as well as Letchford *et al.* (2015) to implement the logarithmical transformation in our regression analysis.

*5.1. Statistical Significance vs. Practical Significance*

Most of the differences aforementioned are not practically significant according to the point estimation results, although there exists some relationship between scientific impact and linguistic complexity. This is consistent with several previous studies, even those with stronger reported relationships between scientific impact and linguistic features. For example, Didegah & Thelwall (2013) found that the length and readability of a paper abstract affected the number of citations of a scientific publication with statistical significance in both the natural and social science areas; however, due to the rather small effective size (i.e., the standardized mean difference between the two groups), they concluded that no practical significance had been found in their results. They focused on features of bibliographic data, whereas we extended that conclusion to full-text features. Letchford, Moat, & Preis (2015) found that publications with shorter titles attracted more citations in a multidisciplinary data set with the 20,000 most cited articles per year (2007–2014) from Scopus. The slope, though statistically significant (p<0.001), suggests that the effect of title length on the logarithm of number of citations is rather low (no more than 0.015); and their Figure 1 (p. 3) suggests that many articles with shorter titles receive relatively fewer citations. In addition, the data set they used might incorporate bias since they excluded papers with a small number of citations. It is difficult for us to observe any practical significance of these findings. Similarly, the results of Sienkiewicz & Altmann (2016, p. 4, Figure 2) failed to show any practical significance in the relationship between linguistic complexity and the number of citations using abstracts for textual analysis.

Statistical significance does not necessarily mean practical significance. How meaningful the difference is between groups should be evaluated by indicators for practical significance (Kirk, 1996; Schneider, 2013). When reviewing the results, one can find that the confidence intervals between groups overlap with each other, even for pairs with statistical significance in Figures 5 and 6. One possible reason is that the group sizes of our Biology sets are relatively large, which makes statistical tests extremely sensitive. Our regression model, however, reinforces our conclusions in that the highest R-square among multiple models is relatively low and the coefficients for independent variables are quite small.

*5.2. The Complexity of Scientific Writing*

Despite the challenges of modeling scientific writing style, these scientific papers usually follow some patterns of linguistic complexity. For example, the average length of sentences in scientific writing is usually greater than 25 words; scientific writings prefer more diverse vocabularies than literary works; and they usually use words comprised of five to ten letters to achieve better comprehension. Modifiers are not heavily used in a text, but some high-impact articles suggest a more frequent adoption of modifiers. These results might also indicate that the complexity of scientific writing is more than just the linguistic complexity, which usually examines the textual or structural complexity of writing. In literacy, linguistic complexity is usually examined with careful manual reading to determine if it is in accordance with the context (e.g., characters, places, and themes). If scientific writing style is to be well modeled, more semantic features or variables need to be developed and employed in the future.

*5.3. Limitations and Future Work*

This study has some limitations. First, we failed to consider the influence of various confounding factors, such as English proficiency (English as a native language, secondary language, foreign language, etc.), size of coauthorship or contributorship, solo or multiauthorship (Cabanac *et al.*, 2014), and readership of a venue (limited readership is likely to translate into limited incoming citations). Second, we have found that many low-impact articles share similar patterns with high-impact ones, which urges us to dig even deeper to discover the hidden factors that may cause failure of those articles. Linguistic complexity is the only linguistic factor considered in this study; some additional factors such as fluency and accuracy should be considered in a future study. Third, we only investigated two disciplines with relatively small samples, especially for Psychology. In addition, the unbalanced group size might also reduce the accuracy of the point estimation results of the high-impact groups in both subjects. Future studies can include more disciplines, more high-impact articles, and more full-text data to draw a more precise and generalizable conclusion.

## ACKNOWLEDGMENTS

The authors would like to thank Evelyn Reynolds for her insightful suggestions for an earlier manuscript of this paper and Yong-Yeol Ahn for his suggestions on the statistical analysis. We are also grateful to our anonymous reviewers and Ludo Waltman for improving the quality of this paper. This work is supported partly by Major Projects of National Social Science Fund of China (No. 16ZAD224), Chinese scholarship Council Project (Student ID: 201606840093), Fujian Provincial Key Laboratory of Information Processing and Intelligent Control (Minjiang University) (No. MJUKF201704), and the Qing Lan Project.

# APPENDIX

We conducted six models for regression analyses, namely, Models 1–6. Suppose that the NC is represented as $y$, while the 12 variables representing linguistic complexity are respectively $x_1$, $x_2$, …, $x_{12}$. The 6 models are:

Model 1: $y = a \sum_{i=1}^{12} x_i^2 + b \sum_{i<>j \; and \; i,j \in [1,12]} x_i x_j + c \sum_{i=1}^{12} x_i + d$

Model 2: $y = a \sum_{i=1}^{12} x_i^2 + b \sum_{i=1}^{12} x_i + c$

Model 3: $lny = a \sum_{i=1}^{12} x_i^2 + b \sum_{i<>j \; and \; i,j \in [1,12]} x_i x_j + c \sum_{i=1}^{12} x_i + d$

Model 4: $lny = a \sum_{i=1}^{12} x_i^2 + b \sum_{i=1}^{12} x_i + c$

Model 5: $y = ln(a \sum_{i=1}^{12} x_i + b)$

Model 6: $lny = a \sum_{i=1}^{12} x_i + b$

Table A1 shows the R-squares of the corresponding models for the Biology data set, in which M1-M6 represent the corresponding models. We can see that most models show very low R-squares, indicating that they cannot interpret the data very well. Model fitting for Psychology is shown in Table A2. For some groups of articles, a collinear relationship exists between independent variables, which lead to no R-square values. The obtained results suggest a weak relationship between linguistic complexity and scientific impact in Psychology as well.

**Table A1. R-squares for regression models in the Biology data set (HS=high-impact group, MS=medium-impact group, LS=low-impact group; the same are used below).**

|    | all   | HS    | MS    | LS    |
|----|-------|-------|-------|-------|
| M1 | 0.028 | 0.089 | 0.019 | 0.030 |
| M2 | 0.021 | 0.049 | 0.010 | 0.021 |
| M3 | 0.045 | 0.105 | 0.017 | 0.036 |
| M4 | 0.032 | 0.030 | 0.010 | 0.026 |
| M5 | 0.018 | 0.022 | 0.008 | 0.019 |
| M6 | 0.029 | 0.031 | 0.008 | 0.023 |

**Table A2. R-squares for regression models in the Psychology data set.**

|    | all   | HS    | MS    | LS    |
|----|-------|-------|-------|-------|
| M1 | 0.026 | -     | 0.351 | 0.062 |
| M2 | 0.007 | -     | 0.000 | 0.027 |
| M3 | 0.050 | -     | 0.348 | 0.066 |
| M4 | 0.016 | -     | 0.000 | 0.031 |
| M5 | 0.007 | 0.562 | 0.000 | 0.024 |
| M6 | 0.016 | 0.312 | 0.000 | 0.028 |